\documentclass[letterpaper]{article} % DO NOT CHANGE THIS
\usepackage{aaai2026}  % DO NOT CHANGE THIS
\usepackage{times}  % DO NOT CHANGE THIS
\usepackage{helvet}  % DO NOT CHANGE THIS
\usepackage{courier}  % DO NOT CHANGE THIS
\usepackage[hyphens]{url}  % DO NOT CHANGE THIS
\usepackage{graphicx} % DO NOT CHANGE THIS
\usepackage{natbib}  % DO NOT CHANGE THIS AND DO NOT ADD ANY OPTIONS TO IT
\usepackage{caption} % DO NOT CHANGE THIS AND DO NOT ADD ANY OPTIONS TO IT
\usepackage{algorithm}
\usepackage{algorithmic}

\usepackage{multirow}
\usepackage{amsmath}
\usepackage{amsfonts}
\usepackage{dsfont}
\usepackage{newfloat}
\usepackage{listings}
\usepackage{pifont}
\DeclareCaptionStyle{ruled}{labelfont=normalfont,labelsep=colon,strut=off} % DO NOT CHANGE THIS
\floatstyle{ruled}
\newfloat{listing}{tb}{lst}{}
\floatname{listing}{Listing}
\title{What Makes Reasoning Invalid: Echo Reflection Mitigation \\ for Large Language Models}
\author {
    Chen He\textsuperscript{\rm 1},
    Xun Jiang\textsuperscript{\rm 1},
    Lei Wang\textsuperscript{\rm 3},
    Hao Yang\textsuperscript{\rm 4},
    Chong Peng\textsuperscript{\rm 4},
    Peng Yan\textsuperscript{\rm 4},
    Fumin Shen\textsuperscript{\rm 1},
    Xing Xu\textsuperscript{\rm 2}
}
\affiliations {
    \textsuperscript{\rm 1}University of Electronic Science and Technology of China\\
    \textsuperscript{\rm 2}School of Computer Science and Technology, Tongji University\\
    \textsuperscript{\rm 3}Salesforce AI Research\\
    \textsuperscript{\rm 4}Meituan\\
}
\usepackage{bibentry}
\begin{document}

\maketitle

\begin{abstract}
% LLM 牛-CoT等技术牛
Large Language Models (LLMs) have demonstrated remarkable performance across a wide range of reasoning tasks. 
%
% Recent advancements, such as Chain of Thoughts (CoTs), reinforcement learning (RL) have further improved LLM performance in complex mathematical reasoning.
Recent methods have further improved LLM performance in complex mathematical reasoning.
However, when extending these methods beyond the domain of mathematical reasoning to tasks involving complex domain-specific knowledge, we observe a consistent failure of LLMs to generate novel insights during the reflection stage. 
%
% Instead of engaging in genuine cognitive refinement, the models tend to mechanically reiterate previous reasoning steps without introducing new information or perspectives, which referred as ``Echo Reflection''.
%
Instead of conducting genuine cognitive refinement, the model tends to  \textit{mechanically reiterate earlier reasoning steps without introducing new information or perspectives, a phenomenon referred to as ``Echo Reflection''}.
We attribute this behavior to two key defects:
% 在生成过程中不受控制的信息流动
(1) Uncontrollable information flow during response generation, which allows premature intermediate thoughts to propagate unchecked and distort final decisions;
(2) Insufficient exploration of internal knowledge during reflection, leading to repeating earlier findings rather than generating new cognitive insights.
Building on these findings, we proposed a novel reinforcement learning method termed \textit{\textbf{A}daptive \textbf{E}ntropy \textbf{P}olicy \textbf{O}ptimization} (\textbf{AEPO}). 
Specifically, the AEPO framework consists of two major components: 
(1) Reflection-aware Information Filtration, which quantifies the cognitive information flow and prevents the final answer from being affected by earlier bad cognitive information;
(2) Adaptive-Entropy Optimization, which dynamically balances exploration and exploitation across different reasoning stages, promoting both reflective diversity and answer correctness.
Extensive experiments demonstrate that AEPO consistently achieves state-of-the-art performance over mainstream reinforcement learning baselines across diverse benchmarks.

\end{abstract}

%     \link{Code}{https://anonymous.4open.science/r/AEPO-7F3A}
%     % \link{Datasets}{https://aaai.org/example/datasets}
%     % \link{Extended version}{https://aaai.org/example/extended-version}

\section{Introduction}
% RL取得了进展
Test-time Scaling (TTS) methodologies have significantly improved the reasoning capabilities of Large Language Models (LLMs) by enabling longer Chain-of-Thought thinking and inducing sophisticated reasoning behaviors.
A pivotal technique driving these improvements is Reinforcement Learning with Verifiable Rewards (RLVR) \cite{shao2024deepseekmath, yu2025dapo, dai2025s, chen2025seed}, where models optimize outputs through RL objectives tied to automated thinking and reflection behavior.

\begin{figure}[!htb]
	\centering
    \includegraphics[width=0.48\textwidth]{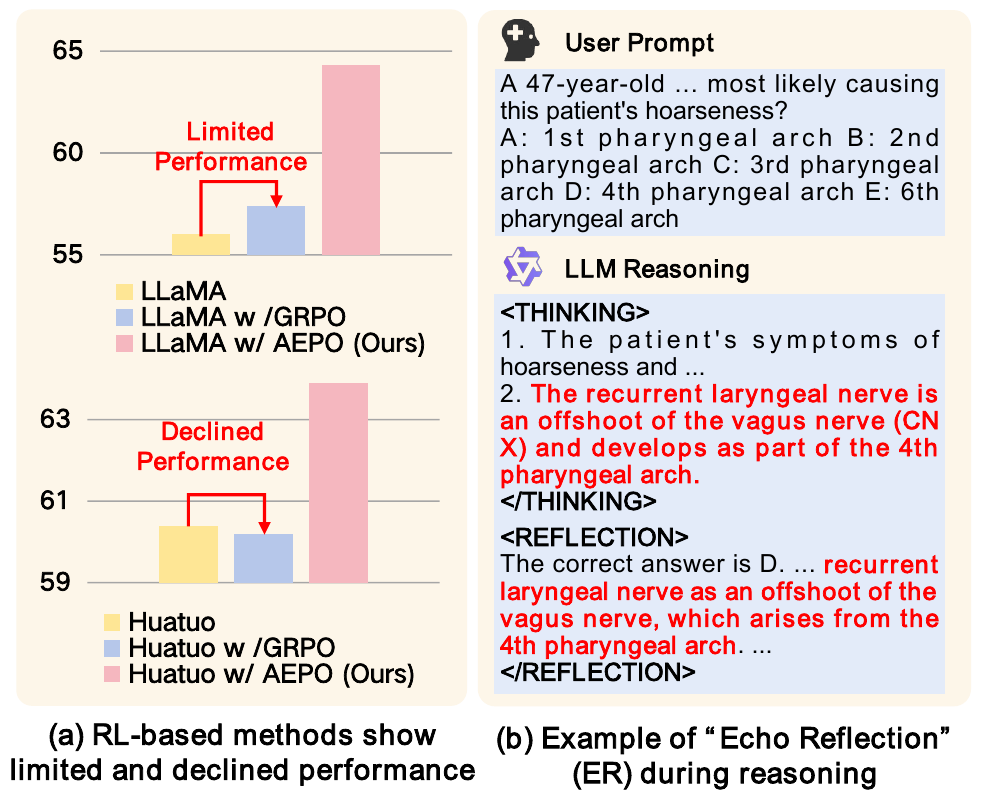}
	\caption{Illustrative examples of: (a) limited or declined performance improvement on original RL-based methods. (b) a typical ``Echo Reflection'' of GRPO-based method reasoning (The echoed wrong contents are highlighted in red).}
	\label{fig:intro}
    \vspace{-5mm}
\end{figure}

% 迁移到其他领域时出现性能下降
Though promising performance is achieved by these methods, we observed a consistent failure when extending these methods into other tasks involving complex domain-specific knowledge. 
% 图1 a说明了这一性能下降
As illustrated in Figure. \ref{fig:intro} (a), GRPO \cite{shao2024deepseekmath} algorithm yields only a marginal improvement of 0.6\% compared with the base model Qwen2.5-7B-Instruct \cite{qwen2}. 
More strikingly, when applied to the Huatuo-O1-70B model \cite{chen2024huatuogpt}, which has been finetuned with extensive domain-specific medical knowledge, the PPO algorithm results in a degradation of performance. 
These findings reveal a fundamental limitation of current reinforcement learning frameworks in enhancing reasoning performance in knowledge-intensive settings.
% 图2 b 展示ER现象
Furthermore, Figure. \ref{fig:intro} (b) presents an example of a failed reasoning attempt by Qwen2.5-7B-Instruct \cite{qwen2}.
In the ``Thinking'' stage, the model erroneously states that ``The recurrent laryngeal nerve is an offshoot of the vagus nerve (CN X) and develops as part of the 4th pharyngeal arch.'' 
In the subsequent ``Reflection'' stage, the model fails to revise this mistake and instead mechanically reiterates the same incorrect reasoning. 
Moreover, due to the narrow scope of the initial ``Thinking'' stage, alternative options are not critically evaluated in the ``Reflection'' stage, leading to shallow analysis and limited knowledge utilization.
This case illustrates that the model failed to fully take advantage of inherent knowledge during reasoning, thereby performing few genuine cognitive updates during reflection.
% 提出ER
We refer to the phenomenon of a model merely repeating earlier reasoning during the reflection stage without generating new insights as \textit{Echo Reflection (ER)}. 
We attribute this failure to two main factors observed in the ``Reflection'' stage:
(1) Uncontrolled propagation of prior errors, where incorrect intermediate steps are preserved;
(2) Insufficient exploration of domain-relevant internal knowledge, which limits the model’s ability to critically assess and revise its initial output. 
Building on this, we propose a novel reinforcement learning algorithm, termed \textit{\textbf{A}daptive \textbf{E}ntropy \textbf{P}olicy \textbf{O}ptimization (\textbf{AEPO})}, designed to mitigate the Echo Reflection problem, thereby enhancing its reasoning capabilities in complex task settings. 

\begin{figure*}[!htb]
	\centering
    \includegraphics[width=0.90\textwidth]{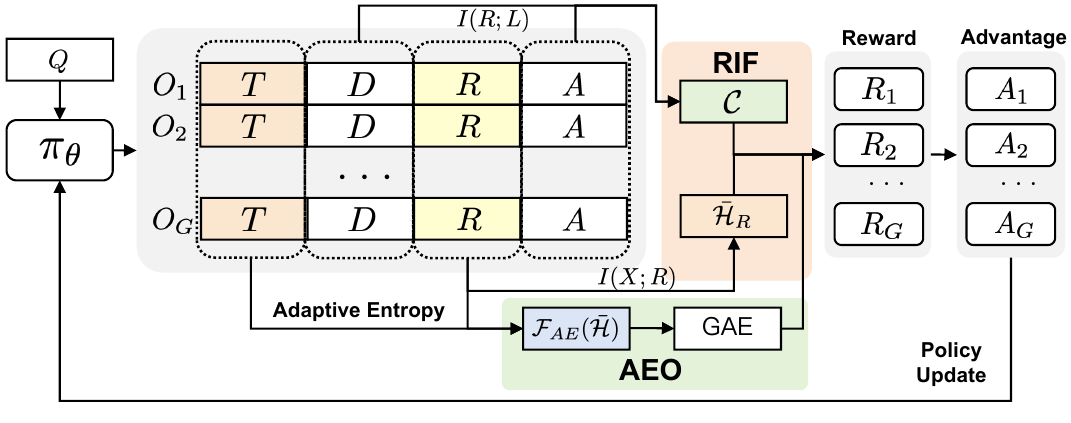}
	\caption{The overall framework of the proposed AEPO algorithm. It consists of two key components: 1) Reflection-aware Information Filtration (RIF), which leverages Information Bottleneck theory to constrain the flow of cognitive information; (2) Adaptive Entropy Optimization (AEO), which balances the behaviors of LLM exploration and exploration.}
	\label{fig:fram}
    % \vspace{-4mm}
\end{figure*}

Specifically, as depicted in Figure \ref{fig:fram}, the AEPO framework consists of two key components: 
% 起名强调不是整个answer的熵而是粒度更细的熵的计算
(1) Reflection-aware Information Filtration (RIF), inspired by Information Bottleneck (IB) theory \cite{tishby2000information}, aims to preserve task-relevant cognitive signals while suppressing misleading intermediate information during reasoning. This module helps regulate the internal flow of information to prevent early-stage errors from contaminating final predictions.
% 整体的算法，强调Adaptive
(2) Adaptive-Entropy Optimization (AEO), which dynamically regulates the policy entropy conditioned on exploration and exploitation balance.

Our contribution can be summarized as follows:

\begin{itemize}
  % 我们识别了EM现象并且揭示了其与步骤级别的策略熵的关系
  \item We identify and characterize the ``Echo Reflection'' phenomenon, shedding light on a previously underexplored failure mode in LLM reasoning over complex, knowledge-intensive tasks.
  % 
  % \item We analyze the root causes of ER from an information-theoretic perspective and propose Reflection-aware Information Filtration, a module grounded in Information Bottleneck theory that constrains the flow of cognitive information during reasoning. 
  \item We take a closer look at the ER from  an information-theoretic perspective and propose Reflection-aware Information Filtration, a module grounded in Information Bottleneck theory that constrains the flow of cognitive information during reasoning. 
  \item We introduce the AEPO algorithm, which enables the model to balance exploration and exploitation during reinforcement learning, encouraging meaningful revisions and deeper knowledge utilization during reflection.
\end{itemize}

\section{Related Works}
\noindent\textbf{Reinforcement Learning in LLMs.}
Reinforcement learning from human feedback (RLHF) \cite{ouyang2022training} has become a key paradigm for aligning large language models with human preferences. Recent advances \cite{schulman2017proximal,shao2024deepseekmath,ahmadian2024back} have extended RLHF beyond instruction-following, introducing verifiable reward functions to better incentivize complex reasoning. OpenAI’s o1 \cite{jaech2024openai} was the first to demonstrate that RL can effectively elicit high-level reasoning abilities in large-scale LLMs. Building on this, models such as DeepSeek R1 \cite{guo2025deepseek}, Qwen QwQ \cite{qwen2}, and Kimi k1.5 \cite{team2025kimi} have sought to match or surpass o1’s performance. 
Concurrently, a series of studies, including Open-Reasoner-Zero \cite{hu2025open}, SimpleRL-Zoo \cite{zeng2025simplerl}, and Logic-RL, \cite{xie2025logic} have explored direct RL-based fine-tuning of base models, omitting the need for an intermediate supervised fine-tuning phase. Other approaches, such as Light-R1 \cite{wen2025light} and DeepScaler \cite{meng2023deepscaler}, propose specially crafted cold-start datasets to encourage fine-grained, step-wise reasoning during early training. The SRPO \cite{zhang2025srpo} framework further combines cold-start strategies with GRPO \cite{shao2024deepseekmath} to enhance deep reasoning development.
In parallel, complementary methods such as VAPO \cite{yue2025vapo}, DAPO \cite{yu2025dapo}, and S-GRPO \cite{dai2025s} aim to refine the GRPO framework by improving reward formulation and advantage estimation, thereby more effectively promoting complex reasoning behaviors in LLMs.

\noindent\textbf{Policy Entropy in Reinforcement Learning.}
Stemmed in information theory, entropy has long been a core principle in reinforcement learning (RL). 
Classical entropy-regularized reinforcement learning (ERL) \cite{ziebart2008maximum, toussaint2009robot} frameworks are often coupled with the soft Bellman update \cite{schulman2017equivalence}, discouraging overly deterministic policies. A canonical example of this approach is the Soft Actor-Critic (SAC) algorithm \cite{haarnoja2018soft}, which effectively constrains policy updates by maintaining proximity to a reference distribution.
Recent work has extended entropy-based methods to LLMs training. ETPO \cite{wen2024entropy} applies token-level ERL to enhance LLM performance in interactive tasks, while EP-PRM \cite{zhang2024entropy} incorporates entropy into process-level reward modeling. Beyond regularization, \citeauthor{cui2025entropy} reveal that entropy dynamics reflect the covariance between action probabilities and logit shifts, offering a tool to prevent policy collapse. \citeauthor{cheng2025reasoning} shows that higher-entropy reasoning chains can lead to more effective inference by generating longer responses.
More abstractly, entropy has been used to model uncertainty at the semantic level. SEED-GRPO \cite{chen2025seed} introduces prompt-level semantic entropy to capture ambiguity in instructions, and \citeauthor{wang2025beyond} identify high-entropy tokens as key ``forks'' during generation.

\begin{figure}[!htb]
	\centering
    \includegraphics[width=0.45\textwidth]{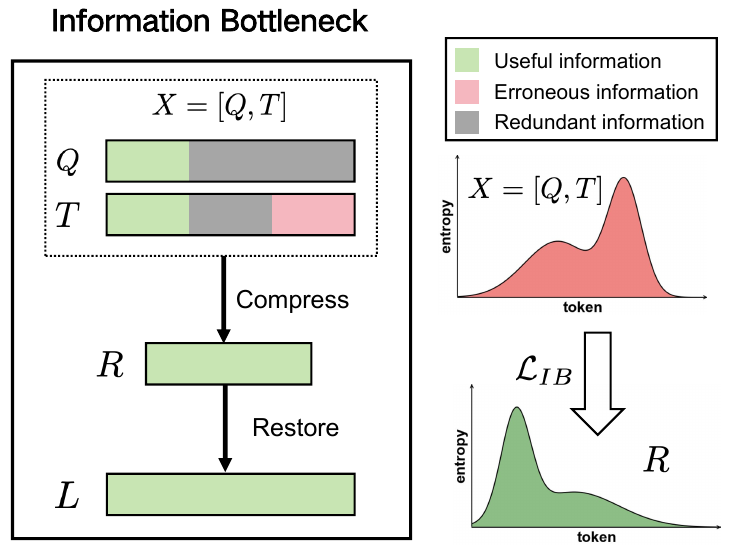}
	\caption{Information Bottleneck adopted in Reflection-aware Information Filtration module. The Information Bottleneck aims to suppress erroneous information and redundant information flow (left). By minimizing $\mathcal{L}_{IB}$, the token entropy is reduced (right).}
	\label{fig:rif}
    % \vspace{-4mm}
\end{figure}

\section{Proposed Method}

\subsection{Preliminary}
% Let $\mathcal{D}$ denote the training dataset, where each data point consists of a question–answer pair $\{Q, L\} \in \mathcal{D}$. Given a large language model (LLM) policy $\pi_{\theta}$, $\mathcal{V}$ as vocabulary, the goal is to generate a response $Y = {y_1, \cdots, y_{|Y|}}$ that effectively addresses the input question $Q$. 
Let $\mathcal{D}$ denote the training dataset, where each data point consists of a question–answer pair $\{Q, L\} \in \mathcal{D}$. 
To optimize the large language model (LLM) policy $\pi_{\theta}$, given $Q$, we sample a group of outputs $\mathcal{O} = \{O_i\}^{G}_{i=1}$ to estimate the advantage. Note that $O_i = \{o_1,\cdots,o_{\vert O_i\vert}\}$ is a complete response sequence sequence consisting of tokens.

We adopt a four-step response format to elicit structured reasoning from the model. Specifically, the model is prompted to sequentially: (1) think about the question (thinking-stage), (2) propose an initial response (draft-stage), (3) perform self-reflection (reflection-stage), and (4) produce a final answer (answer-stage). 
To simplify notation, we also refer to the sampled output $O_i$ as being composed of four semantically coherent segments: $O_i = [T, D, R, A]$, where each segment corresponds to a distinct stage of reasoning. Note that while $O_i$ denotes the full token sequence generated by the model, $T$, $D$, $R$, and $A$ are contiguous token subsequences that represent the outputs of each respective step. 
The prompt used for the four-step response format is detailed in the \textit{supplementary material}.

\subsection{Reflection-aware Information Filtration}
To prevent LLMs from failing to accurately identify errors in previous reasoning during the reflection process, the Reflection-aware Information Filtration module aims to optimize the model's reflection by constraining the information flow, suppressing the retention of erroneous or redundant information, and promoting the transmission and retention of accurate and useful information.

\noindent \textbf{Information Bottleneck.}
% IB 理论和符号对应
The Information Bottleneck theory \cite{tishby2000information} introduces an intermediate variable $Z$, whose objective is to compress the input as much as possible while retaining sufficient information to recover the target output. 

As shown in Figure~\ref{fig:rif}, in the context of LLM reasoning, $Q$ along with $T$ together serve as input. We use $X = [Q, T]$ to represent them.
The reflection $R$ serves as the intermediate representation, and the groundtruth answer $L$ as the output.

IB theory formulates this trade-off using mutual information (MI) to quantify statistical dependence. 
The MI between $X$ and $R$ is denoted as $I(X;R)$, which measures how much information about $X$ is preserved in $R$. 
A higher mutual Information indicates that more of the $X$ can be reconstructed for $R$, whereas a lower value suggests that $R$ is a compressed, less redundant encoding of $X$. 
However, minimizing $I(X;R)$ alone can lead to loss of task-relevant information. To address this, IB theory simultaneously maximizes $I(R;L)$, which is the MI between $L$ and $R$. 
In this manner, the IB theory ensures the retained information in $R$ remains sufficient for accurately predicting $L$.

To this end, the optimization objective is:
\begin{equation}
\mathcal{L}_{IB} = I(X; R) - \beta I(R; L),
\end{equation}
where $\beta$ is a hyper-parameter. 

Under this formulation, minimizing $I(X; R)$ encourages the reflection to filter out redundant or erroneous content, reducing mechanical repetition of earlier reasoning. Concurrently, maximizing $I(R; L)$ promotes the retention of task-relevant information within the reflection, thereby alleviating the ``Echo Reflection'' phenomenon.

\noindent \textbf{Proxy Metrics.}
While the IB objective provides a principled framework for guiding information flow during reflection, directly computing mutual information terms $I(X; R)$ and $I(R; L)$ is intractable in high-dimensional generative models such as LLMs \cite{alemi2017deep, shwartz2022opening}. 
% However, mutual information terms are intractable to compute directly for high-dimensional discrete sequences such as language \cite{alemi2017deep, shwartz2022opening}. 
To this end, we propose the Proxy Metrics for an efficient estimate of the mutual information.

Specifically, the token-level entropy of the reflection step is denoted as:

\begin{equation}
    \label{eq:entropy}
\mathcal{H}_{o_i} = -\sum_{v\in\mathcal{V}}\pi_{\theta}(v|q,o_{<i})\mathrm{log}\pi_{\theta}(v|q,o_{<i}).
\end{equation}
Hence, the average policy entropy $\bar{\mathcal{H}}_R$ over the reflection stage is:
\begin{equation}
\bar{\mathcal{H}}_R = {1 \over {|R|}}\sum_{y_i \in R}\mathcal{H}_{o_i}.
\end{equation}

The $\bar{\mathcal{H}}_R$, measures the information capacity of the reflection received from the thinking. 
Lower entropy implies higher predictability of the reflection given the initial reasoning, indicating stronger dependence (i.e., higher mutual information). 
Thus, we interpret higher entropy as a signal of reduced dependence on $T$, aligning with the goal of minimizing $I(X,R)$.

% Then define a contribution indicator $\mathcal{C}$ to evaluate the utility of the reflection in improving the final answer. 
% It is computed based on the model’s initial response $D$, final answer $A$ and groundtruth label $L$.

To estimate $I(R; L)$, we define a contribution indicator $\mathcal{C}$, which measures how much the reflection improves the likelihood of producing a correct answer. 
Higher $\mathcal{C}$ reflects greater task-relevant information retained in $R$, corresponding to higher $I(R; L)$:
\begin{equation}
    I(R; L) = \mathcal{C} = 
\begin{cases}
0.4, & \text{if $D = L$ and $ A = L$} \\
0.6, & \text{if $D \neq L$ and $ A = L$} \\
0, & \text{if $D \neq L$ and $ A \neq L$} \\
-0.3, & \text{if $D = L$ and $ A \neq L$} .
\end{cases}
\end{equation}

\subsection{Adaptive Entropy Optimization}
% ER 的另一个原因
Another key factor that leads to the ER phenomenon is the insufficient exploration of internal knowledge during reasoning. To tackle this problem, we adopt the Adaptive Entropy Optimization mechanism to balance the model's exploration and exploitation.

% \noindent \textbf{DAPO Algorithm.}
% \citeauthor{yu2025dapo} imporved the GRPO \cite{shao2024deepseekmath} and presented DAPO. Specifically, the DAPO samples g group of outputs $\{o_i\}^{G}_{i=1}$ for $Q$ and the groundtruth $L$ and optimizes the policy via the following objective:

% \begin{equation}
% \begin{aligned}
%     \small
% \mathcal{L}_{\text{DAPO}}(\theta) 
% = &\ \mathbb{E}_{(Q, L) \sim \mathcal{D}, \{o_i\}_{i=1}^{G} \sim \pi_{\theta_{\text{old}}}(\cdot \mid Q)} \\ & \Bigg[ 
%  \frac{1}{\sum_{i=1}^{G} |o_i|} 
%    \sum_{i=1}^{G} \sum_{t=1}^{|o_i|} 
%    \min \Big( r_{i,t}(\theta) \hat{A}_{i,t},\   \\
% & \quad \text{clip}\left( r_{i,t}(\theta), 1 - \epsilon_{\text{low}}, 1 + \epsilon_{\text{high}} \right) \hat{A}_{i,t} \Big) \Bigg].
% \end{aligned}
% \end{equation}

% where
% \begin{equation}
% \begin{aligned}
%  r_{i,t}(\theta) = \frac{\pi_\theta(o_{i,t} \mid Q, o_{i,<t})}{\pi_{\theta_{\text{old}}}(o_{i,t} \mid Q, o_{i,<t})} \\
%  \hat{A}_{i,t} = \frac{R_i - \text{mean}(\{R_i\}_{i=1}^G)}{\text{std}(\{R_i\}_{i=1}^G)}.
% \end{aligned}
% \end{equation}

\noindent \textbf{Adaptive Entropy.} 
% We build upon the standard DAPO framework, which estimates token-level advantages.
To quantify the degree of exploration at each stage of reasoning, we compute the average token-level entropy for the thinking-stage and reflection stage with Equation~\ref{eq:entropy}, yielding $\mathcal{\bar{H}}_T$ and $\mathcal{\bar{H}}_T$, respectively.

A common practice in reinforcement learning is to inject entropy into the Bellman backup to encourage exploration. Following this, one might directly maximize $\mathcal{\bar{H}}_T$ and $\mathcal{\bar{H}}_T$. However, excessive entropy can lead to incoherent reasoning, while recent findings~\cite{agarwal2025unreasonable} show that minimizing entropy may improve precision.

To balance these extremes, we aim to maintain entropy near a task-suitable target. Prior work~\cite{wang2025beyond} observes a phase transition around entropy $\mathcal{H}^{*} = 0.67$ , above which reasoning becomes more effective. Based on this, we adopt an adaptive entropy mechanism as follows:

\begin{equation}
\begin{aligned}
\mathcal{F}_{AE}(\mathcal{\bar{H}}) = - \Vert \mathcal{\bar{H}} - \mathcal{H}^{*}\Vert_2,
\end{aligned}
\end{equation}
where $\mathcal{\bar{H}} \in \{\mathcal{\bar{H}}_T, \mathcal{\bar{H}}_R\}$. 
This adaptive mechanism guides both thinking and reflection stages to balance exploration and exploitation, respectively.

\noindent \textbf{Gated Adaptive Entropy.}
While exploration fosters diversity, it also introduces the risk of incorrect reasoning. 
% In our framework, such errors in the thinking stage are implicitly suppressed by the Information Bottleneck mechanism. However, the reflection stage lacks similar constraints and is more susceptible to propagating errors.
To mitigate this, we adopt a Gated Adaptive Entropy (GAE) to ensure entropy rewards for reflection are gated by final correctness—only applied when the output is correct. 
%The resulting optimization objective becomes:
\begin{equation}
\begin{aligned}
\mathcal{F}_{GAE} = \sum_{\mathcal{\bar{H}} \in \{\mathcal{\bar{H}}_T, \mathcal{\bar{H}}_R\}}\mathcal{F}_{AE}(\mathcal{\bar{H}}) * \mathds{1}_{\text{correct}}.
\end{aligned}
\end{equation}
Hence the total objective for output $O_i$ is:
\begin{equation}
\begin{aligned}
\mathcal{L}(O_i) = min (\mathcal{L}_{IB} - \mathcal{F}_{GAE}).
\end{aligned}
\end{equation}

\noindent \textbf{Formulation of AEPO.}

To facilitate the calculation, following previous workss \cite{yu2025dapo,guo2025deepseek}, we use token-level policy gradient loss, which is denoted as: 

\begin{equation}
\begin{aligned}
    \tiny
\mathcal{L}_{\text{AEPO}}(\theta) 
= &\ \mathbb{E}_{(Q, L) \sim \mathcal{D}, \{O_i\}_{i=1}^{G} \sim \pi_{\theta_{\text{old}}}(\cdot \mid Q)} \Bigg[ 
 \frac{1}{\sum_{i=1}^{G} |O_i|}  \\ &
   \sum_{i=1}^{G} \Big[ \mathcal{L}(O_i) + \sum_{t=1}^{|O_i|} 
   \min \big( r_{i,t}(\theta) \hat{A}_{i,t},\   \\
& \quad \text{clip} \left( r_{i,t}(\theta), 1 - \epsilon_{\text{low}}, 1 + \epsilon_{\text{high}} \right) \hat{A}_{i,t} \big) \Big] \Bigg],
\end{aligned}
\end{equation}
where
\begin{equation}
\begin{aligned}
 r_{i,t}(\theta) = \frac{\pi_\theta(o_{i,t} \mid Q, O_{i,<t})}{\pi_{\theta_{\text{old}}}(O_{i,t} \mid Q, o_{i,<t})} \\
 \hat{A}_{i,t} = \frac{R_i - \text{mean}(\{R_i\}_{i=1}^G)}{\text{std}(\{R_i\}_{i=1}^G)}.
\end{aligned}
\end{equation}

% The full algorithm can be found in Algorithm \ref{alg:aepo}:

% \begin{algorithm}[!thb]
% 	\caption{Algorithm for AEPO.}
% 	\label{alg:aepo}

% 	\textbf{Input}: policy model $\pi_\theta$; dataset $\mathcal{D}$; hyperparameters $\epsilon_{\text{low}}, \epsilon_{\text{low}}$; training steps $S$. 

%         \textbf{Output}: $\pi_\theta$.

%         \begin{algorithmic}[1]

%         \FOR{$\text{step} = 1,\cdots,S$}
%         \STATE Sample a minibatch $\mathcal{D}_m$ form $\mathcal{D}$.
%         \STATE Update the old policy model $\pi_{\theta_{old}} \leftarrow \pi_\theta$.
%         \STATE Sample $G$ outputs $\{O_i\}^G_{i=1} \sim \pi_{\theta_{old}}$ for $Q \in \mathcal{D}_m$.
%         \STATE Compute reward $r_i$ for $O_i$.
%         \STATE Compute $\mathcal{L}({O_i}) = min(\mathcal{L}_{IB}-\mathcal{F}_{GAE})$ for $O_i$.
%         \STATE Compute advantage $\hat{A}_{i,t}$ for each token for $O_i$.
%         \STATE Update $\pi_{\theta}$ by maximizing the AEPO objective.
%         \ENDFOR
%         \end{algorithmic}
% \end{algorithm} 

% Hence the objective is:
% \begin{equation}
% \begin{aligned}
% \mathcal{L}_{\text{AEO}} = \mathcal{L}_{\text{DAPO}} + \mathcal{F}_{AE}(\mathcal{\bar{H}}_T) + \mathcal{F}_{AE}(\mathcal{\bar{H}}_R) * \mathbf{1}_{correct}
% \end{aligned}
% \end{equation}

% Combining this with the IB-based reward yields the final training loss:
% \begin{equation}
% \begin{aligned}
% \mathcal{L}_{\text{AEPO}} = \mathcal{L}_{\text{IB}} + \mathcal{L}_{\text{AEO}}.
% \end{aligned}
% \end{equation}

\section{Experiments}

\subsection{Experimental Setup}

\noindent \textbf{Datasets.} 
We evaluate the proposed AEPO framework on both in-distribution (I.D.) and out-of-distribution (O.O.D.) datasets to assess its generalization and robustness.

The I.D. datasets include:
(1) MedQA \cite{jin2021disease}, which contains 10,178 training and 1,273 testing QA pairs from professional medical board exams; 
(2) MedMcQA \cite{pal2022medmcqa}, which contains 182,822 training and 4,183 testing QA pairs from real-world medical entrance exams.
% Specially, \citeauthor{tang2025medagentsbench} further split both MedQA and MedMcQA into normal split and hard split, we simultaneously report the results on the two splits.

The O.O.D datasets include: 
(1) MMLU-Pro \cite{wang2024mmlu}, which is a multi-task language understanding benchmark comprising 12,000 questions spanning diverse academic domains;
(2) GPQA \cite{rein2024gpqa}, which is a subject-specific QA dataset covering biology, physics, and chemistry domains. 
(3) MATH-500, which contains a subset of 500 problems from the MATH benchmark that OpenAI \cite{lightman2023let}.
(4) AIME 24, which contains problems from the American Invitational Mathematics Examination (AIME) 2024.
(5) AMC 23, which is a mathematical dataset with 40 problems.

\begin{table}[htb!]
    \centering
    \small
\begin{tabular}{c|c|c}
\hline
% \multirow{2}{*}{Model} & \multirow{2}{*}{\#Params} & \multirow{2}{*}{\#Params}         \\ 
Model & \#Params & Accuracy \\\hline \hline
                    %    &                           & \multicolumn{1}{c|}{Normal} \\ \hline \hline
Mistral                & 7B                        & {48.2}       \\
Yi-1.5                 & 9B                        & {50.8}      \\ \hline
LLaMA-3.1                & 8B                        & {58.7}       \\
LLaMA-3.1 + CoT          & 8B                        & {57.7}             \\
LLaMA-3.1 + GRPO         & 8B                        & {63.8}            \\
LLaMA-3.1 + DAPO         & 8B                        & {62.9}             \\ \hline
Qwen2.5                & 7B                        & {57.0}             \\
Qwen2.5 + CoT          & 7B                        & {57.7}            \\
Qwen2.5 + GRPO         & 7B                        & {58.4}            \\
Qwen2.5 + DAPO         & 7B                        & {57.1}            \\ \hline
AEPO-L (Ours)          & 8B                        & {\textbf{68.5}}            \\
AEPO-Q (Ours)          & 7B                        & {62.5}            \\ \hline
\end{tabular}
\caption{Comparison Results on MedQA Dataset. The AEPO-L and AEPO-Q indicate LLaMA-3.1 w/ AEPO and Qwen2.5 w/ AEPO, respectively.}
\label{tab:main_medqa}
\end{table}

\begin{table}[htb!]
    \centering
    \small
\begin{tabular}{c|c|c}
\hline
% \multirow{2}{*}{Model} & \multirow{2}{*}{\#Params} & \multicolumn{2}{c}{Splits}         \\ \cline{3-4} 
%                        &                           & \multicolumn{1}{c|}{Normal} & Hard \\ \hline \hline
Model & \#Params & Accuracy \\\hline \hline
Mistral                & 7B                        & {44.6}       \\
Yi-1.5                 & 9B                        & {48.7}       \\ \hline
LLaMA-3.1                & 8B                        & {56.0}      \\
LLaMA-3.1 + CoT          & 8B                        & {56.6}            \\
LLaMA-3.1 + GRPO         & 8B                        & {57.4}         \\
LLaMA-3.1 + DAPO         & 8B                        & {59.8}         \\ \hline
Qwen2.5                & 7B                        & {55.6}           \\
Qwen2.5 + CoT          & 7B                        & {56.4}            \\
Qwen2.5 + GRPO         & 7B                        & {56.2}          \\
Qwen2.5 + DAPO         & 7B                        & {57.4}         \\ \hline
AEPO-L (Ours)          & 8B                        & {\textbf{64.3}}            \\
AEPO-Q (Ours)          & 7B                        & {61.6}            \\ \hline
\end{tabular}
\caption{Comparison Results on MedMcQA Dataset. The AEPO-L and AEPO-Q indicate LLaMA-3.1 w/ AEPO and Qwen2.5 w/ AEPO, respectively.}
\label{tab:main_medmcqa}
\end{table}

\noindent \textbf{Training Details.}
All models are trained on 4 Nvidia A6000 GPU, using the EasyR1 framework \cite{zheng2025easyr1}. We adopt Qwen2.5-7B-Instruct \cite{qwen2} (denoted as Qwen-7B) and LLaMA3-8B-Instruct \cite{llama3} (denoted as LLaMA-8B) as base models for most experiments. 
Training is conducted with bfloat16 precision using the AdamW \cite{loshchilov2017decoupled} optimizer, a constant learning rate of $1 \times 10^{-6}$, and a linear warm-up over the first 10 steps. 
Each batch contains 64 prompts, with 5 responses sampled per prompt. 
For language model decoding, we set the temperature to $1.0$ and top-p to $0.99$. 
Following DAPO \cite{yu2025dapo}, we use clipped policy ratios with $\epsilon_{low} = 0.2$ and $\epsilon_{high} = 0.28$. The $\beta$ is $1$.

\noindent \textbf{Inference Details.}
We report QA accuracy on both I.D. and O.O.D. datasets. 
In I.D. setting, models are trained on the training split and evaluated on the corresponding test split. 
In O.O.D. setting, we directly evaluate the I.D.-trained models on unseen O.O.D. datasets without further adaptation. 
For inference stability, the decoding temperature is fixed at 0 during inference.

\subsection{Overall Comparison Results}
\begin{table}[htb!]
\centering
\small
\begin{tabular}{c|c|cc|cc|cc}
\hline
\multirow{2}{*}{No.} & \multirow{2}{*}{RIF}       & \multicolumn{2}{c|}{AEO}   & \multicolumn{2}{c|}{MedQA} & \multicolumn{2}{c}{MedMcQA} \\ \cline{3-8} 
                     &    & \multicolumn{1}{c|}{AE} & GAE & Llama        & Qwen        & LLaMA         & Qwen        \\ \hline
0 & -  & \multicolumn{1}{c|}{-}  & -                & 58.7            & 57.0           & 56.0             & 55.6           \\
1 & \ding{52} & \multicolumn{1}{c|}{-}  & -         & 62.3            & 58.4           & 56.6             & 56.8           \\
2 & -  & \multicolumn{1}{c|}{\ding{52}} & -         & 63.7            & 58.6           & 57.9             & 57.8           \\
3 & -  & \multicolumn{1}{c|}{-}  & \ding{52}        & 59.4            & 57.8           & 57.4             & 57.4           \\ \hline
4 & \ding{52} & \multicolumn{1}{c|}{\ding{52}} & -  & 65.8            & 60.7           & 59.3             & 58.8           \\
5 & -  & \multicolumn{1}{c|}{\ding{52}} & \ding{52} & 66.0            & 61.2           & 61.5             & 59.2           \\
6 & \ding{52} & \multicolumn{1}{c|}{-}  & \ding{52} & 65.4            & 60.2           & 59.4             & 58.6           \\ \hline
7 & \ding{52} & \multicolumn{1}{c|}{\ding{52}} & \ding{52} & \textbf{68.5}            & \textbf{62.5}           & \textbf{64.3}             & \textbf{61.6}           \\ \hline
\end{tabular}
\caption{Ablation Study on MedQA and MedMcQA dataset.}
\label{tab:ablation}
\end{table}

\begin{table}[htb!]
\centering
\small
\begin{tabular}{c|cc}
\hline
Dataset     & MedQA & MedMcQA \\ \hline
Baseline    & 72.6  & 60.4    \\
w/ GRPO        & 72.4  & 57.3    \\
w/ DAPO        &  71.2     &61.7         \\ \hline
w/ AEPO (Ours) &  \textbf{75.4}     &\textbf{63.9}        \\ \hline
\end{tabular}
\caption{Comparison Results on Domain Specific Finetune.}
\label{tab:huatuo}
\end{table}

\begin{table}[htb!]
\centering
\small
\begin{tabular}{c|cc}
\hline
Dataset     & MedQA & MedMcQA \\ \hline
Baseline    & 58.4  & 56.5    \\
w/ GRPO        & 59.2  & 58.2    \\
w/ DAPO        & 62.4      &  60.1       \\ \hline
w/ AEPO (Ours) &  \textbf{68.5}     &\textbf{63.7}         \\ \hline
\end{tabular}
\caption{Comparison Results on Additional Prompts.}
\label{tab:prompts}
\end{table}

\noindent \textbf{Comparisons On In-Distribution Dataset.} 
To evaluate the effectiveness of our proposed AEPO method in enhancing LLM reasoning in knowledge-intensive question answering, we compare AEPO with mainstream reinforcement learning with verifiable reward (RLVR) approaches across multiple LLM backbones.
Answer accuracy is reported on both the MedQA and MedMcQA benchmarks, as shown in Table~\ref{tab:main_medqa} and Table~\ref{tab:main_medmcqa}, respectively.
From the experimental results, we can observe that:
(1) AEPO consistently outperforms existing RLVR baselines, including GRPO and DAPO, across both the standard and hard splits, demonstrating its effectiveness in mitigating the Echoed Reasoning phenomenon during multi-stage inference;
(2) Specifically, our AEPO outperforms the DAPO by 5.4\% and 5.6\% on Qwen-7B and LLaMA-8B, respectively. 
These remarkable results demonstrate that our proposed AEPO method effectively enhances reasoning by suppressing incorrect reasoning in the thinking stage and promotes more accurate final answers.
(3) Moreover, AEPO achieves superior performance across both MedQA and MedMcQA, indicating its stronger capacity to explore and leverage the internal knowledge embedded within large language models.

\begin{figure}[!htb]
	\centering
    \includegraphics[width=0.45\textwidth]{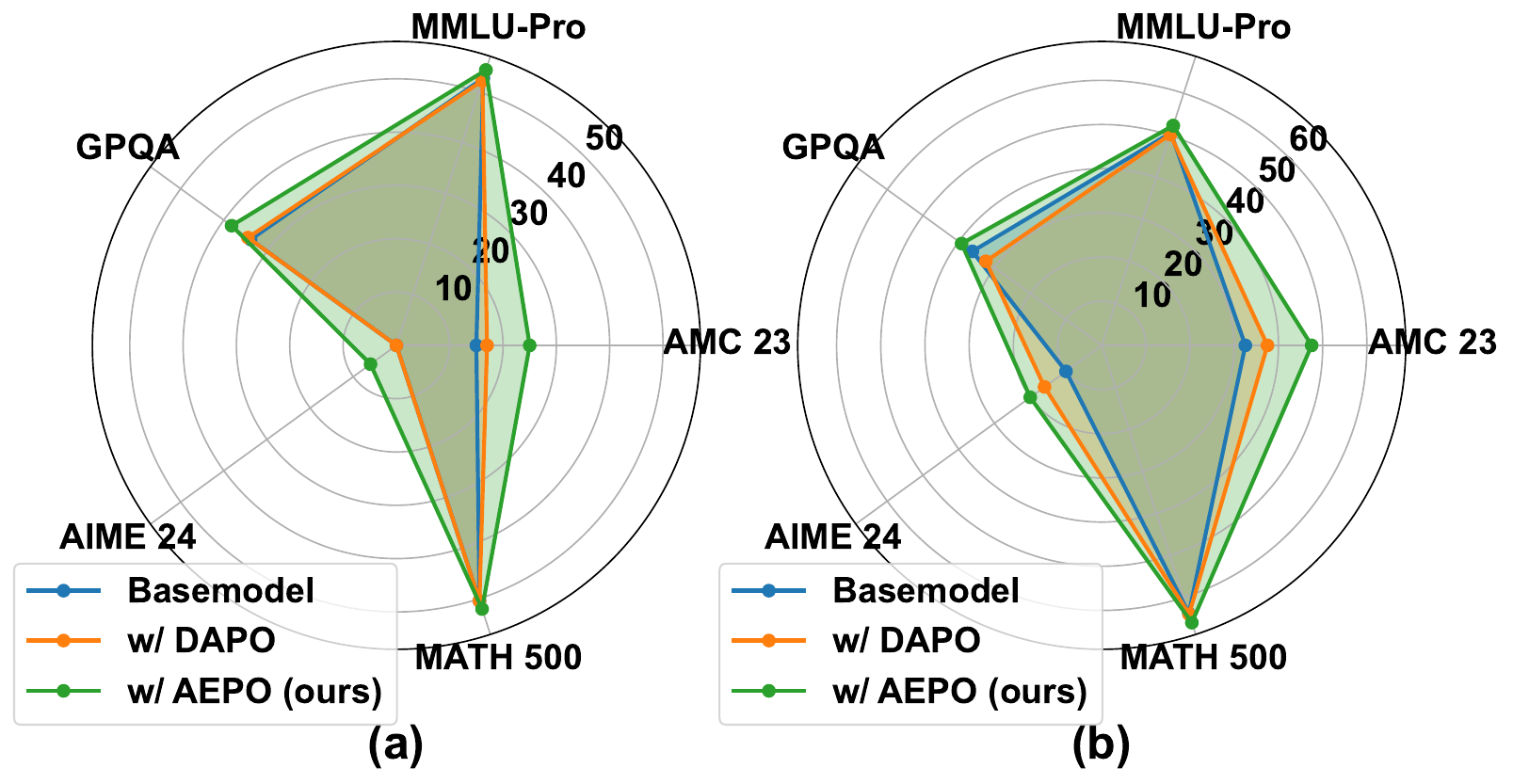}
	\caption{Visualization of performances on O.O.D. datasets.}
	\label{fig:radar}
\end{figure}

\noindent \textbf{Comparisons On Out-of-Distribution Dataset.}
To further evaluate the generalization capability of our model in handling O.O.D. problems, we conduct additional experiments on unseen datasets and compare our method with state-of-the-art (SOTA) baselines.
Specifically, as shown in Figure~\ref{fig:radar}, subfigure (a) is the experimental results of LLaMA-8B trained on the MedMcQA dataset, and subfigure (b) is the experimental results of Qwen-7B trained on MedQA. We directly validate the O.O.D. without additional training.
The results reveal the following insights:
(1) Our proposed AEPO method significantly outperforms existing SOTA methods on both domains, demonstrating stronger generalization to unfamiliar problem types.
(2) We attribute this improved generalization to the Adaptive Entropy Optimization module, which enables deeper and more efficient exploration of the solution space while avoiding policy collapse.
(3) Moreover, AEPO effectively filters out irrelevant or misleading information during open-ended reasoning, enabling the model to maintain coherent and accurate inference paths even under high uncertainty.

\subsection{Further Analysis}
\noindent \textbf{Ablation Study.} 
As shown in Table~\ref{tab:ablation}, we conduct ablation studies on both MedQA and MedMcQA to examine the contribution of each component in our AEPO framework. From the experimental results, the following observations can be made:
(1) The comparison between setting No.0 and No.1 demonstrates the effectiveness of the proposed RIF module. By constraining the cognitive information flow and suppressing the influence of erroneous or redundant intermediate reasoning, RIF significantly improves the model’s ability to generate correct answers.
(2) The performance gap between settings No.1 and No.6 highlights the impact of Adaptive Entropy Optimization. By dynamically regulating the policy entropy in the thinking stage, AEO enables deeper yet controlled exploration. When combined with RIF, this leads to notably improved reasoning performance.
(3) The comparison between settings No.4 and No.7 reveals the importance of the Correctness-Gated Entropy Reward mechanism. By conditioning entropy-based exploration rewards on the correctness of the final answer, this component helps ensure that exploratory behaviors contribute meaningfully to reasoning quality.

\noindent \textbf{Comparisons On Finetuned Model.}
To comprehensively assess the effectiveness of the proposed AEPO method in leveraging domain-specific knowledge, we further conduct experiments on HuatuoGPT-O1-8B \cite{chen2024huatuogpt}, which is finetuned specifically for the medical domain.
As shown in Table~\ref{tab:huatuo}, the following observations can be made:
(1) Incorporating domain-specific knowledge leads to significant performance gains for HuatuoGPT-O1-8B over the base models. However, existing RLVR methods show limited additional improvements, indicating their inability to effectively utilize domain-specific information.
(2) In contrast, our AEPO method consistently outperforms both HuatuoGPT variants by 3.0\% and 4.2\% on MedQA, demonstrating its superior capacity to exploit internal knowledge. We attribute this to the AEO module, which facilitates deeper and more controlled exploration of the solution space while mitigating the risk of policy collapse.

\noindent \textbf{Analysis on Additional Prompts.}
To further demonstrate the robustness of our proposed AEPO method across different Chain-of-Thought (CoT) prompting strategies, we conduct additional experiments using alternative prompts.
In addition to the four-stage reasoning prompt introduced in our main framework, we also adopt an R1-style CoT prompt following \citet{guo2025deepseek}.
Details of the prompt format are provided in the \textit{supplementary material}.
Experimental results show that under the R1-style CoT setting, AEPO significantly outperforms the GRPO baseline.
It highlights the prompt-robustness of AEPO and its strong transferability across different reasoning paradigms.

\begin{figure}[!htb]
	\centering
    \includegraphics[width=0.45\textwidth]{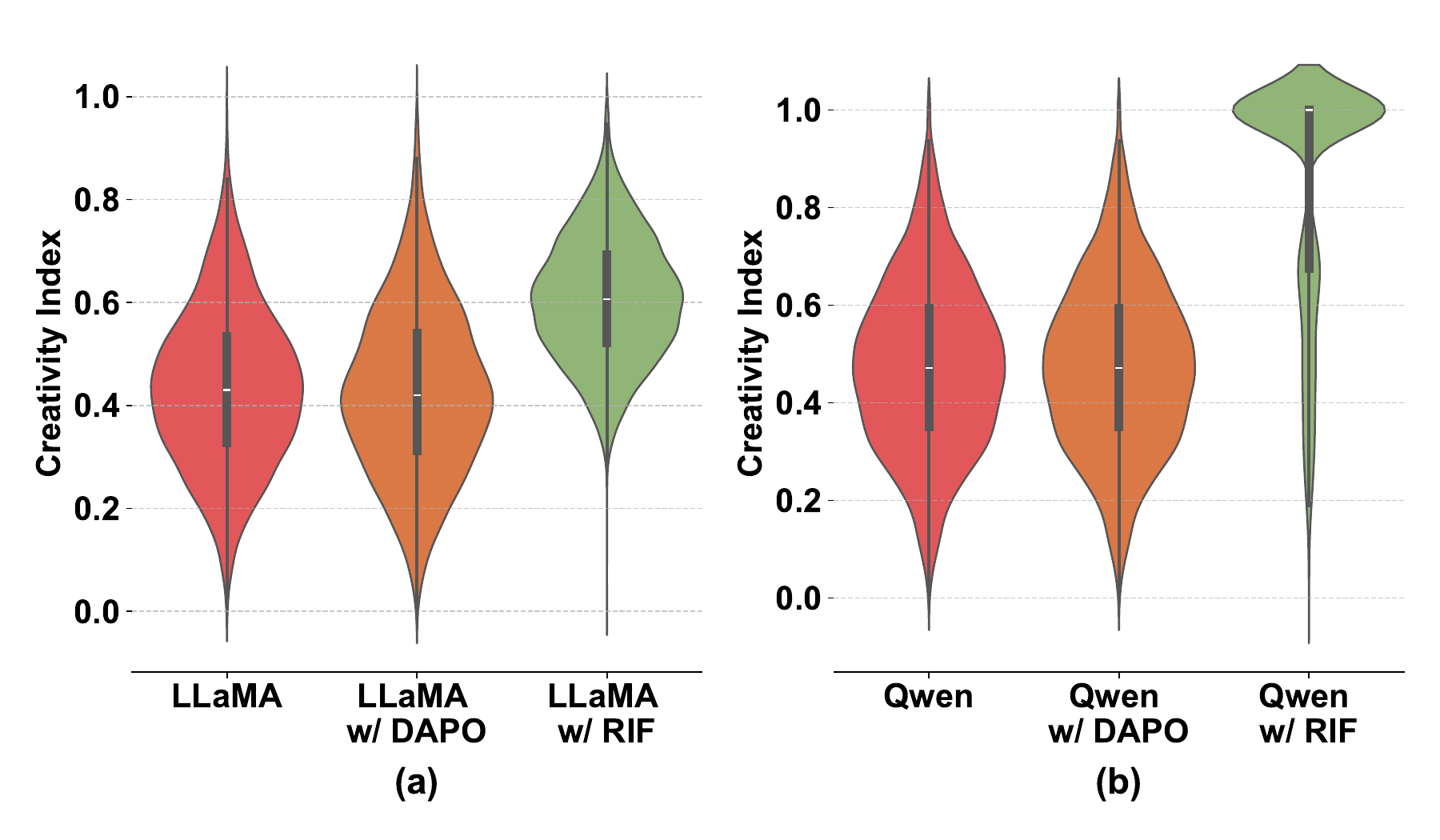}
	\caption{Violin plot of creativity index on MedMcQA.}
	\label{fig:CI}
\end{figure}

\begin{figure}[!htb]
	\centering
    \includegraphics[width=0.45\textwidth]{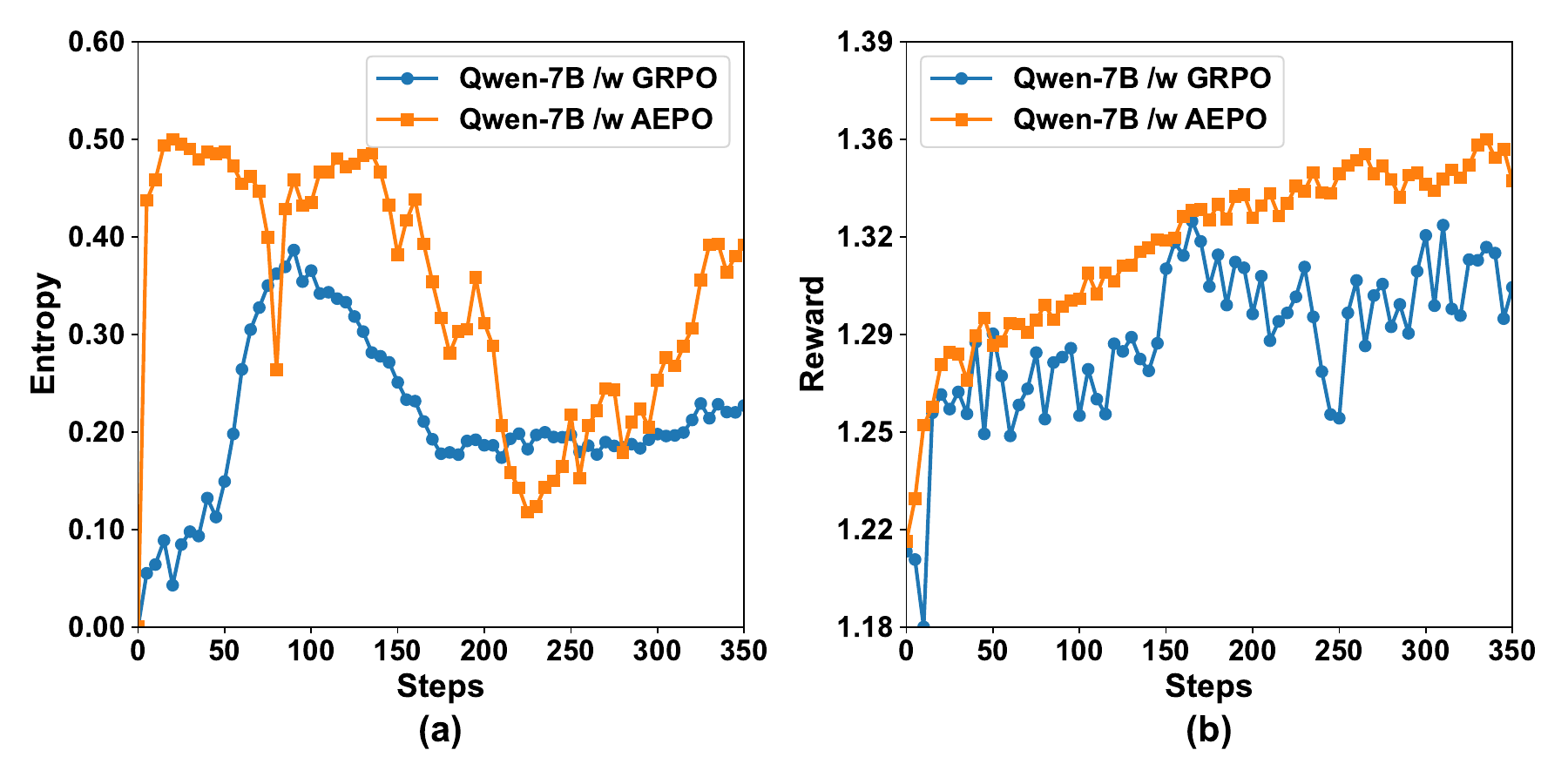}
	\caption{Visualization of training dynamics.}
	\label{fig:training}
\end{figure}

\begin{figure*}[!htb]
	\centering
    \includegraphics[width=0.95\textwidth]{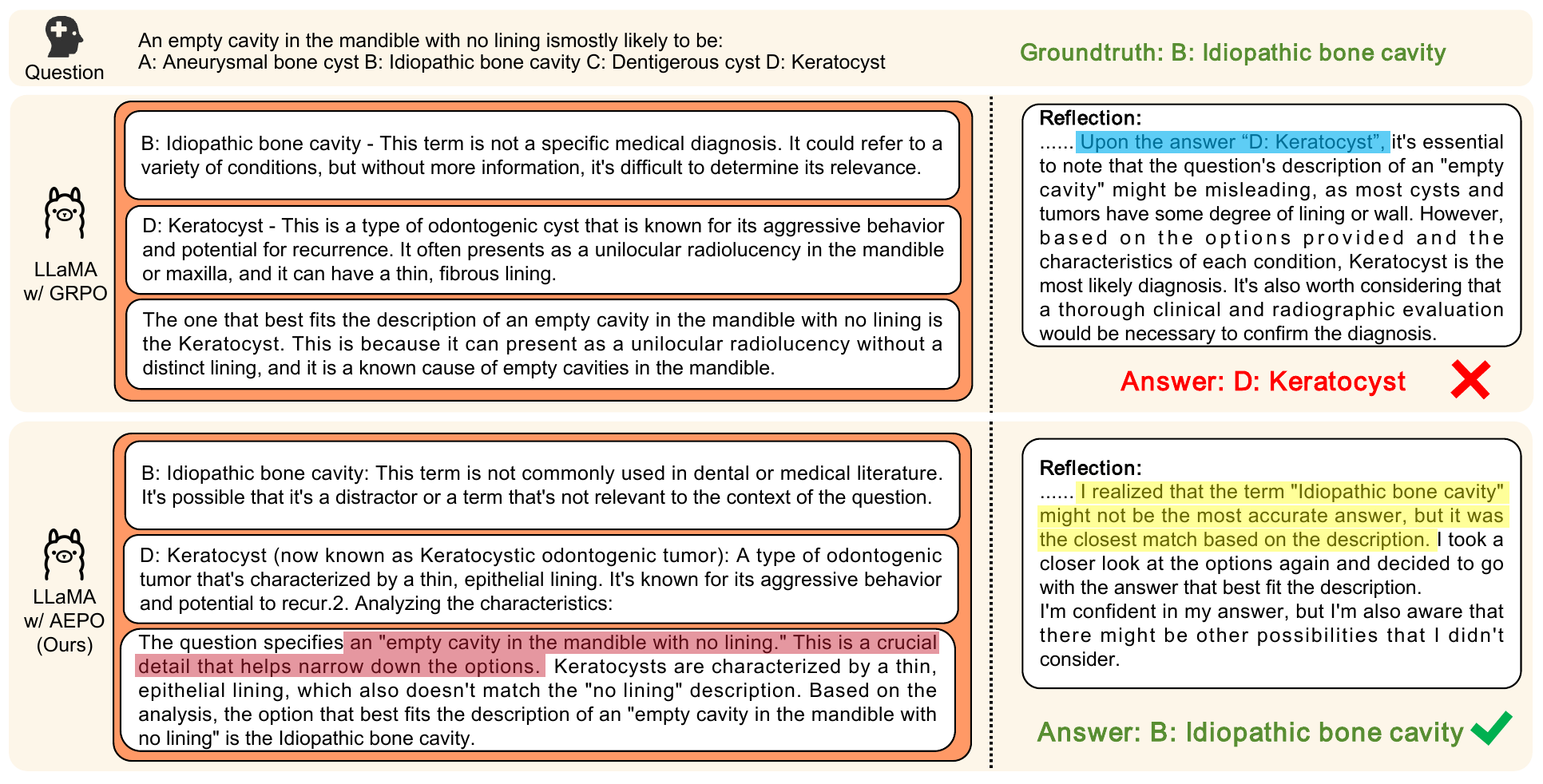}
	\caption{Quality visualization analysis. It presents responses of LLaMA with GRPO (LLaMA w/GRPO) and our proposed AEPO method (LLaMA w/AEPO). The left side of the dashed line illustrates the content of thee model's thinking-stage, while the right side displays the content of the reflection-stage and the final answer. Key contents are highlighted in different colors.}
	\label{fig:qa}
\end{figure*}

\noindent \textbf{Analysis on Creativity Index.} 
To highlight the influence of our proposed RIF module, we visualize the Creativity Index \cite{Quantifying} of LLaMA models on the MedMcQA dataset under different RL strategies using violin plots. The CI measures the creativity of model outputs, where higher values indicate more novel generations, while lower values suggest a tendency to copy from the reference corpus.
In our setting, we construct the reference corpus by concatenating all the input questions with the model’s response of the thinking-stage, and then evaluate the CI of the corresponding reflection stage. 
%This setup allows us to quantify how much new information the model contributes during reflection, beyond what was already produced during initial reasoning.
The results reveal that both LLaMA-8B w/ RIF and Qwen-7B w/ RIF achieve substantially higher Creativity Index scores compared to their counterparts with DAPO. This suggests that the RIF module effectively suppresses the influence of prior thinking content during reflection, thereby mitigating the echo reflection phenomenon and encouraging more original reflective reasoning.

\noindent \textbf{Analysis on RL Training Dynamics.} 
To further illustrate the advantages of the proposed AEPO algorithm, we visualize the training dynamics of policy entropy and reward during training for two methods: GRPO and AEPO.
As shown in Figure \ref{fig:training} (a), the GRPO exhibits a rapid decay in policy entropy, leading to premature convergence to a suboptimal policy.
However, our proposed AEPO algorithm effectively mitigates the entropy collapse problem commonly observed in reinforcement learning while also avoiding the instability caused by excessive entropy.
Figure \ref{fig:training} (b) further supports this claim: while DAPO quickly converges after entropy drops, AEPO is able to maintain effective exploration throughout training.
We attribute this behavior to the AEO module, which strikes a dynamic balance between exploration and exploitation.
This balance enables AEPO to sustain high-quality exploration of domain-specific knowledge, ultimately leading to improved reasoning performance.

\noindent \textbf{Qualitative Analysis.} 
As shown in Figure~\ref{fig:qa}, we present a qualitative analysis that compares the question answer results from the MedMcQA dataset.
For clarity, we only visualize the critical reasoning contents leading to divergent decisions between LLaMA w/ GRPO and our LLaMA w/ AEPO. 
The visualization reveals that, during the thinking stage, LLaMA w/ AEPO conducts a more refined analysis of the options. 
Notably, when evaluating the correct option B, LLaMA w/ AEPO successfully connects ``bone cavity'' and ``no lining'' (highlighted with a red rectangle), enabling a correct judgment. This capability is largely attributed to our proposed AEO module, which dynamically adjusts entropy to balance exploration and exploitation.
Moreover, during the reflection phase, LLaMA w/ AEPO demonstrates enhanced metacognitive awareness. For instance, it explicitly acknowledges: ``Idiopathic bone cavity might not be the most accurate answer'' (highlighted with a blue rectangle). This indicates that the model maintains critical awareness of terminology, avoiding cognitive fixation.
In contrast, LLaMA w/ GRPO makes an incorrect judgment early in reasoning and, during reflection, fails to correct the error, instead reiterating its flawed reasoning (highlighted with a yellow rectangle). This highlights the vulnerability of conventional frameworks to error propagation and entrenchment. Our Reflection-aware Information Filtering module mitigates such Echo Reflection by selectively filtering and reconstructing intermediate reasoning information. More qualitative analysis on the medical domain and other domains can be found in the \textit{supplementary material}.

\section{Conclusion}
In this paper, we identified the Echo Reflection (ER) phenomenon, where language models fail to perform meaningful cognitive updates during reflection. 
Through step-wise policy entropy analysis, we found that low entropy suppresses the retrieval and utilization of inherent knowledge, leading to ER.
To address this, we proposed Adaptive Entropy Policy Optimization (AEPO), which enhances reasoning by jointly optimizing two components: Reflection-aware Information Filtration, guided by Information Bottleneck theory, and Adaptive Entropy Optimization, formulated via dynamic entropy adjustment. 
Together, they effectively mitigate ER by promoting more informative and adaptive reasoning processes.
In the future, we will explore additional strategies to further improve reasoning effectiveness.

\bibliography{aaai2026}
\end{document}